\documentclass[10pt,twocolumn,letterpaper]{article}

\usepackage[pagenumbers]{cvpr}              
\usepackage{comment}

\usepackage[dvipsnames]{xcolor}
\definecolor{cvprblue}{rgb}{0.21,0.49,0.74}
\usepackage[pagebackref,breaklinks,colorlinks,citecolor=cvprblue]{hyperref}

\usepackage{algorithm}
\usepackage{algorithmic}
\def\paperID{13086} 
\def\confName{CVPR}
\def\confYear{2024}

\title{Meta Co-Training: Two Views are Better than One}

\author{Jay C.~Rothenberger \hspace*{1em} Dimitrios I.~Diochnos\\
University of Oklahoma\\
{\tt\small \{jay.c.rothenberger, diochnos\}@ou.edu}
}

\usepackage{color} 
\definecolor{RoyalBlue}{cmyk}{1, 0.50, 0, 0}
\definecolor{BeamerBlue}{rgb}{0.2, 0.202, 0.698}
\definecolor{ForestGreen}{cmyk}{0.864, 0.0, 0.429, 0.396}
\definecolor{Brown}{cmyk}{0.0,0.692,0.925,0.529}
\definecolor{MyGreen}{cmyk}{0.95, 0.05, 0.95, 0.05}
\definecolor{MyPurple1}{rgb}{0.45,0.353,0.963}
\definecolor{MyPurple2}{rgb}{0.63,0.4,0.63}
\definecolor{MyYellow}{rgb}{0.901,0.547,0.0}
\definecolor{MyRed}{rgb}{0.9,0.0,0.0}
\definecolor{MyStylishGreen}{rgb}{0.328,0.601,0.169}
\usepackage{booktabs} 
\usepackage{amsthm}
\usepackage{multirow}
\usepackage{xspace}
\newcommand{\abs}[1]{\left\lvert #1\right\rvert}
\newcommand{\norm}[1]{\left\lVert#1\right\rVert}
\newcommand{\set}[1]{\ensuremath{\{#1\}}}
\usepackage{amsmath}
\DeclareMathOperator*{\argmax}{arg\,max}
\DeclareMathOperator*{\argmin}{arg\,min}

\newcommand{\RR}{\ensuremath{\mathbb R}\xspace}

\newcommand{\sOO}[1]{\ensuremath{%
    \ifthenelse{\equal{#1}{}}{\widetilde{\mathcal{O}}}{\widetilde{\mathcal{O}} \left( #1 \right)\xspace}%
  }}

\newcommand{\XX}{\ensuremath{\mathcal{X}}\xspace} 
\newcommand{\YY}{\ensuremath{\mathcal{Y}}\xspace} 

\newcommand{\risk}[2]{\ensuremath{R_{#1}\left(#2\right)}\xspace}
\newcommand{\emprisk}[2]{\ensuremath{\widehat{R}_{#1}\left(#2\right)}\xspace}

\newcommand{\charfunc}[1]{\ensuremath{\mathbf{1}\left\{#1\right\}}\xspace}

\newcommand{\DD}{\ensuremath{D}\xspace} 

\newcommand{\expectedsub}[2]{\ensuremath{\mathbf{E}_{#1}\left[#2\right]}\xspace}

\newif\ifcomments
\commentstrue

\newcommand{\Dnote}[1]{\ifcomments{\color{blue}[D: #1]}\xspace\fi}
\newcommand{\Jnote}[1]{\ifcomments{\color{MyRed}[J: #1]}\xspace\fi}

\usepackage{nicefrac}

\usepackage{wrapfig}
\usepackage{framed}

\newcommand{\ourparagraph}[1]{\paragraph{#1.}}

\begin{document}

\maketitle

\label{sec:abstract}
\begin{abstract}
In many critical computer vision scenarios unlabeled data is plentiful, but labels are scarce and difficult to obtain.  As a result, semi-supervised learning which leverages unlabeled data to boost the performance of supervised classifiers have received significant attention in recent literature.  One representative class of semi-supervised algorithms are \emph{co-training} algorithms.  Co-training algorithms leverage two different models which have access to different independent and sufficient representations or ``views'' of the data to jointly make better predictions. Each  of these models creates pseudo-labels on unlabeled points which are used to improve the other model.  
We show that in the common case where independent views are not available, we can construct such views inexpensively using pre-trained models. Co-training on the constructed views yields a performance improvement over any of the individual views we construct and performance comparable with recent approaches in semi-supervised learning.  
We present \emph{Meta Co-Training}, a novel semi-supervised learning algorithm,
which has two advantages over co-training:
\emph{(i)} learning is more robust when there is large discrepancy between the information content of the different views, and
\emph{(ii)} does not require retraining from scratch on each iteration.
Our method achieves new state-of-the-art performance on ImageNet-10\% achieving a $\sim4.7\%$ reduction in error rate over prior work. Our method also outperforms prior semi-supervised work on several other fine-grained image classification datasets.

\end{abstract}


\section{Introduction}
\label{sec:introduction}

In many critical machine learning scenarios we have access to a large amount of unlabeled data, and relatively few labeled data points for a given task.  For standard computer vision (CV) tasks there are large well-known and open source datasets with hundreds of millions or even billions of unlabeled images~\cite{LAION5B,LAION400M}.  In contrast, labeled data is usually an order of magnitude more scarce and otherwise expensive to obtain requiring many human-hours to generate.  In this context, \emph{semi-supervised learning (SSL)} methods are useful as they rely on training more performant models, using small amounts of labeled data and large amounts of unlabeled data.

Not unrelated to, and not to be confused with the idea of \emph{\textbf{semi}-supervised learning}, is \emph{\textbf{self}-supervised learning}.\footnote{There is also an SSL technique 
called \emph{self-training}. Recognizing the unfortunate similarity of these three terms, SSL will always mean \emph{semi-supervised learning} in this text and we will not abbreviate the other terms.} 
In self-supervised learning a model is trained with an objective that does not require a label that is not evident from the data itself.  Self-supervised learning was popularized by the BERT model~\cite{SSL:ST-BERT} for generating word embeddings for natural language processing (NLP) tasks.  BERT generates its embeddings by solving the \emph{pretext} task of masked word prediction.  Pretext tasks present unsupervised objectives that are not directly related to any particular supervised objective we might want to solve, but rather are solved in the hope of learning a suitable representation to more easily solve a downstream task.  
These pretext learners are often referred to as \emph{foundation models}.  
Several pretext tasks have been proposed for CV to train associated foundation models that solve them~\cite{SSL:ST-MAE,SSL:ST-SimCLR,SSL:ST-EsViT,SSL:ST-SwAV,SSL:ST-DINOv2,SSL:ST-MoCo}.  
The learned representations for images are often much smaller than the images themselves. As a consequence, this yields the additional benefit of reduced computational cost when using the learned representations.

SSL algorithms~\cite{SSL:ST:PL, SSL:co-training, SSL:co-training:deep, SSL:NST, SSL:MPL, SSL:FixMatch, SSL:UDA, SSL:MixMatch} involve generating pseudo-labels to serve as weak supervision on unlabeled data and then learning from those pseudo-labels.  This weak supervision can either be used to perform \emph{consistency regularization}, or \emph{entropy minimization}, or both.  Consistency regularization is to enforce that different augmented versions of the same instance are assigned the same label.  Entropy minimization is to enforce that all instances are assigned a label with high confidence.  Typically consistency regularization is achieved by minimizing a consistency loss between pairs of examples that ought to have the same label~\cite{SSL:UDA, SSL:MixMatch, SSL:FixMatch, SSL:ReMixMatch}, and entropy minimization is achieved by iterative re-training~\cite{SSL:ST:PL, SSL:co-training, SSL:co-training:deep} or sharpening labels~\cite{SSL:MPL, SSL:MixMatch, SSL:FixMatch, SSL:ReMixMatch}.

One particular class of SSL algorithms are co-training~\cite{SSL:co-training} algorithms in which two different ``views'' of the data must be obtained or constructed, and then two different models must be trained and re-trained iteratively.  These two views yield models which capture different patterns in their input and thus, informally speaking, are more likely to fail independently.  This independence of failure is leveraged so that each model can provide useful labels to the other.  Standard benchmark problems in machine learning do not present two views of the problem to be leveraged for co-training.  If one hopes to apply co-training, one needs to construct two views from the single view that they have access to.  
We show that constructing views which satisfy the conditions necessary for co-training is relatively simple.  Training different models on the two constructed views can boost performance.  Unfortunately, the classical \emph{co-training} algorithm (CT) fails to utilize pseudo-labels effectively on benchmark tasks.

We propose a novel SSL method which more effectively leverages pseudo-labels. We call our algorithm \emph{Meta} Co-Training (MCT), because it leverages two views to produce good pseudo-labels as part of a bi-level optimization. In MCT each model is trained to provide better pseudo-labels to the other model, given only its view of the data and the performance of the other model on a labeled set.  Similarly to CT, our approach utilizes multiple views to train models which are diverse.  These models have an advantage when teaching (and learning from) each-other because they will learn different patterns from their differing input data.  We show that our approach provides an improvement over both CT and the current state-of-the-art (SoTA) method~\cite{REACT} on the ImageNet-10\% dataset, as well as it establishes new SoTA few-shot performance on several other fine-grained image classification tasks.

\ourparagraph{Summary of Contributions}
Our contributions are 
summarized as follows.
\begin{enumerate}
    \item We propose \emph{Meta Co-Training}, an SSL algorithm which 
    appears to be more robust than co-training and 
does not require iterative re-training from initialization.
    
    \item We establish SoTA top-1 ImageNet-10\%~\cite{Dataset:ImageNet} accuracy and other few-shot benchmark classification tasks.
    
    \item We make our implementation publicly available for others to use and build upon.
    \url{https://github.com/JayRothenberger/Meta-Co-Training}
\end{enumerate}

\ourparagraph{Outline of the Paper}
In Section~\ref{sec:background} we establish notation and provide background information necessary to understand our method.
In Section~\ref{sec:methods} we describe our proposed method, meta co-training.  In Section~\ref{sec:experiments} we present our experimental findings on ImageNet~\cite{Dataset:ImageNet} and discuss how our method compares to relevant baseline approaches.  We perform additional experiments on Flowers102~\cite{Dataset:Flowers102}, Food101~\cite{Dataset:Food101}, FGVCAircraft~\cite{Dataset:FGVCAircraft}, iNaturalist~\cite{Dataset:iNaturalist}, and iNaturalist 2021~\cite{Dataset:iNat2021} datasets. Additional details such as hyperparameters are provided in Appendix~\ref{sec:appendix:hyperparameters}.  In Section~\ref{sec:related} we discuss related work with emphasis on semi-supervised methods, including co-training methods.  In Section~\ref{sec:conclusion} we conclude with a summary and directions for future work.

\section{Background}\label{sec:background}
\ourparagraph{Notation}
The models that we learn are functions of the form $f \colon \XX \rightarrow \Delta^{\vert \YY \vert}$ where $\Delta^{\vert \YY \vert}$ is the $\vert \YY \vert$-dimensional unit simplex. 
We use $f(x)|_j$ to denote the $j$-th value of $f$ in the output.  When we want to explicitly refer to a network $f$ parameterized by $\theta$ (e.g.,~in deep neural networks), we write $f_{\theta}$.
%
For the entirety of this text $\ell$ will refer to the cross-entropy loss which is defined as $\ell\left(y, f_{\theta}(x)\right) = -\sum_{\xi\in\YY}\charfunc{\xi = y}\log\left(f_{\theta}(x)|_{\xi}\right)$. 
We use $\eta$ as the \emph{learning rate} and $T$ as the \emph{maximum number of steps (or updates)} in a gradient-based optimization procedure.
Finally, we use $\charfunc{\mathcal{A}}$ as an indicator function of an event $\mathcal{A}$; that is, $\charfunc{\mathcal{A}}$ is equal to 1 when $\mathcal{A}$ holds, otherwise 0.

\ourparagraph{Semi-Supervised Learning}
We are interested in situations where we have a large pool of unlabeled instances $U$ as well as a small portion $L$ of them that is labeled. 
That is, 
the learning algorithm has access to a dataset $S = L \cup U$, such that $L = \left\{(x_i, y_i)\right\}_{i=1}^{m}$ and $U = \left\{x_j\right\}_{j=1}^{u}$.  For $L$ the instance part is denoted as $X_L$ and the label part is denoted as $Y_L$. In this setting, \emph{semi-supervised learning (SSL)} methods attempt to first learn an initial model $f_{init}$ using the labeled data $L$ via a supervised learning algorithm, and in sequence, an attempt is being made in order to harness the information that is hidden in the unlabeled set $U$, so that a better model $f$ can be learnt.

\ourparagraph{View Construction}\label{sec:view-construction}
A view, in the context of co-training methods, is a subset of the input features. 
A view 
typically has two properties:
\emph{(i)} it is sufficient for predicting the desired quantity, and 
\emph{(ii)} it is conditionally independent of other views given the label.  Previous methods of view construction for co-training include manual feature subsetting~\cite{SSL:co-training:view-construction:ActiveDisagreement}, automatic feature subsetting~\cite{SSL:co-training:view-construction:AutoPart}, 
random feature subsetting~\cite{SSL:co-training:view-construction:RandomSubset}, random subspace selection~\cite{SSL:co-training:view-construction:RandomSubsetEM}, and adversarial examples~\cite{SSL:co-training:deep}.  
We use self-supervised learning to construct views for our experiments.
However, if one \emph{has access} to a dataset where the instances have two different views, then one can still obtain useful representations with this approach using the same self-supervised model or none at all.  

While our approach is more widely applicable, it is particularly well-suited to computer vision.  There are a plethora~\cite{SSL:ST-BYOL,SSL:ST-SimCLR,SSL:ST-SwAV,SSL:ST-EsViT,SSL:ST-CLIP,SSL:ST-DINOv2,SSL:ST-MAE} of competitive learned representations for images.  
Choosing foundation model representations as views makes our approach lightweight enough to be feasible with limited hardware resources.  
\section{Proposed Method: Meta Co-Training}\label{sec:methods}\label{sec:MCT}

We propose a novel semi-supervised learning algorithm called \emph{Meta Co-Training (MCT)}, 
which is described below.  MCT operates within a bi-level student-teacher optimization framework.  A student is optimized to replicate the labels its teacher gives to a set of unlabeled points while a teacher is optimized to provide labels to the same unlabeled points such that its student learns to predict well on labeled data.  We call MCT a co-training algorithm because it operates on two views of the data, however unlike the original co-training method \cite{SSL:co-training} we do not train multiple models for each view sequentially.  Each view has exactly one model which acts as both a teacher for the other view and a student for its own view.  That is to say that the model learning on view 1 will take an unlabeled instance from view 1 as input and provide the pseudo-label prediction to the student for the same unlabeled instance in view 2.  The student -- the model learning on view 2 -- will use the pseudo-label to learn to predict similarly.

The lower of the two levels of the optimization is the student optimization (Equation \ref{eq:SPL}).  The student parameters $\theta_S$ are optimized as a function of the teacher parameters $\theta_T$:
\begin{equation}\label{eq:LU}
\mathcal{L}_u\left(\theta_T, \theta_S\right) = \ell\left(\argmax_{\xi} f_{\theta_T}(U)|_{\xi}, f_{\theta_S}(U)\right)
\end{equation}
\begin{equation}\label{eq:SPL}
\theta_S' = \theta_{S}^{\text{PL}}(\theta_T) \in 
\argmin_{\theta_S}\mathcal{L}_u\left(\theta_T, \theta_S\right)\,.
\end{equation}

The teacher optimization (Equation \ref{eq:MCTTprime}) forms the upper level of the optimization:
\begin{equation}\label{eq:LL}
\mathcal{L}_L\left(\theta_S'\right) = \ell\left( Y_L, f_{\theta_S'}(X_L)\right)
\end{equation}
\begin{equation}\label{eq:MCTTprime}
\theta_T' = \argmin_{\theta_T} \displaystyle\mathcal{L}_L\left(\theta_S'\right)\,.
\end{equation}

All together the objective of MCT from the perspective of the current view model as the teacher is
\begin{equation}\label{eq:MCT:loss}
\min_{\theta_T} \mathcal{L}_u\left(\theta_T, \theta_S\right) + \mathcal{L}_L\left(\theta_S'\right)\,.
\end{equation}

It is cumbersome to speak of teachers and students when each model is both.  We give the objective of MCT in Equation \ref{eq:MCT:joint} as jointly optimizing the following objectives with $f_{\theta_1}$ the model learning on view 1 and $f_{\theta_2}$ the model learning on view 2.
\begin{equation}\label{eq:MCT:joint}
    \min_{\theta_1, \theta_2} \mathcal{L}_u\left(\theta_1, \theta_2\right) + \mathcal{L}_L\left(\theta_2'\right)\ + \mathcal{L}_u\left(\theta_2, \theta_1\right) + \mathcal{L}_L\left(\theta_1'\right)\,
\end{equation}
%

Co-training algorithms are traditionally evaluated using the joint prediction of the constituent models; 
i.e., 
the re-normalized element-wise product of the individual predictions.

\begin{figure}
    \centering
    \includegraphics[width=0.80\columnwidth]{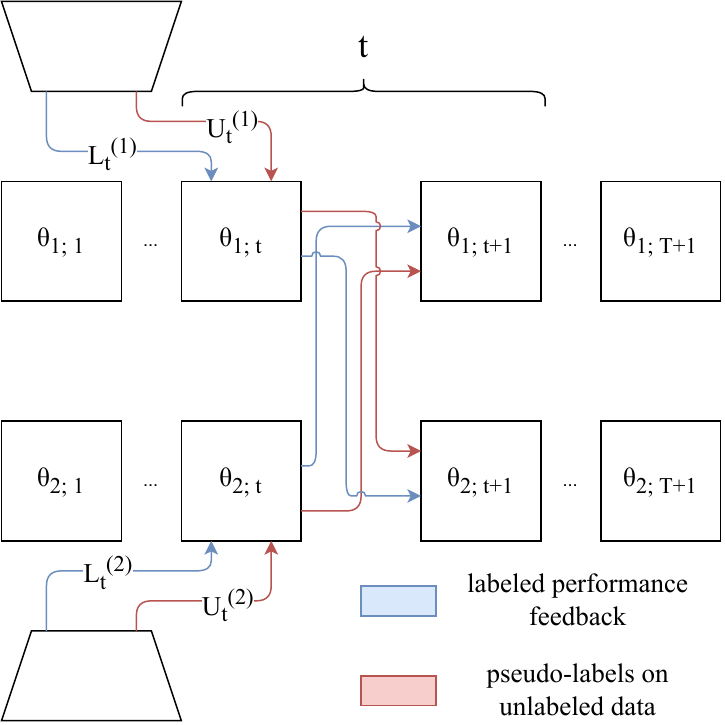}
    \caption{
    At each step $t\in\set{1, \ldots, T}$ of meta co-training the models that 
    correspond to the so-far learnt parameters $\theta_{1; t}$ and $\theta_{2; t}$
    play the role of the student and the teacher simultaneously using batches for their respective views.  Pseudo-labeling occurs on complementary views so that 
    the teacher can provide the student with labels on an unlabeled batch.
    Labeled batches may, or may not, use complementary views as the purpose that they serve is to calculate the risk of the student model on the labeled batch and this result signals the teacher model to update its weights accordingly.
    }
    \label{fig:MCT-Overview}
    \bigskip
\end{figure}

To make tractable the optimization in Equation \ref{eq:MCT:joint} we approximate the student and teacher optimizations using alternating iterations of stochastic gradient descent.  In Algorithm~\ref{alg:MCT} the student updates are in lines 13-14 while the teacher updates are in lines 24-25.  An overview of MCT is provided in Figure~\ref{fig:MCT-Overview}.
At each step $t\in\set{1, \ldots, T}$ of meta co-training the models that 
correspond to the so-far learnt parameters $\theta_{1; t}$ and $\theta_{2; t}$
participate in a student-teacher framework in which each model plays the role of the student for their view and the teacher for the other model \emph{simultaneously}.  
%
%
The representations obtained from the view construction process of Section~\ref{sec:view-construction} form different views that are used by MCT. 

By training two models on two views to collaborate we can take advantage of the aspects of MPL and Co-Training that make each algorithm successful.  Co-training is more effective when the two constituent models fail (predict incorrectly) independently.  During MCT the teacher model will receive positive feedback precisely when it makes a prediction that the other model would not have made which results in an improvement.  MCT trains models in a way which encourages each teacher to express independence while it teaches each student to incorporate the improvement.  Naturally, once the two models predict similarly then the gradient of the student parameters with respect to the teacher's labels is zero and they both cease to improve. This, however, is also when MPL ceases to improve and it is unlikely to have models which are as diverse as those which are trained from independent views. 

\begin{algorithm}[ht]
\caption{\textbf{Meta Co-Training} (Section~\ref{sec:background} has notation)} 
\label{alg:MCT} 
\begin{algorithmic}[1]
\STATE Input: $L^{(1)}, L^{(2)}, U^{(1)}, U^{(2)}, f^{(1)}, f^{(2)}, \theta_{1}, \theta_{2}, T, \ell, \eta$
\FOR{step $t \in \{1 \dots T\}$} 
\STATE // Unlabeled views must be complementary
\STATE $U_t^{(1)} \gets SampleBatch(U^{(1)})$
\STATE $U_t^{(2)} \gets getOtherView(U_t^{(1)}, 2)$
\STATE // Predict soft pseudo-labels
\STATE $\hat{y}^{(1)} \gets f^{(1)}_{\theta_1}(U_t^{(1)})$
\STATE $\hat{y}^{(2)} \gets f^{(2)}_{\theta_2}(U_t^{(2)})$
\STATE // Sample the pseudo-labels from the discrete distribution over the classes
\STATE $PL^{(1)} \sim \hat{y}^{(1)}$ 
\STATE $PL^{(2)} \sim \hat{y}^{(2)}$ 
\STATE // MCT student update applied to both models
\STATE $\theta'_{1} \gets \theta_{1} - \eta \cdot \nabla_{\theta_{1}}\ell(PL^{(2)}, \hat{y}^{(1)})$
\STATE $\theta'_{2} \gets \theta_{2} - \eta \cdot \nabla_{\theta_{2}}\ell(PL^{(1)}, \hat{y}^{(2)})$
\STATE // Sample batches from $L$ for the teacher updates
\STATE $X_t^{(1)}, Y_t^{(1)} \gets SampleBatch(L^{(1)})$
\STATE $X_t^{(2)}, Y_t^{(2)} \gets SampleBatch(L^{(2)})$
\STATE // MCT teacher update applied to both models
\STATE $\hat{y}'^{(1)} \gets f^{(1)}_{\theta'_1}(X_t^{(1)})$ 
\STATE $\hat{y}'^{(2)} \gets f^{(2)}_{\theta'_2}(X_t^{(2)})$ 
\STATE $h^{(1)} \gets \nabla_{\theta'_{2}}\ell(Y_t^{(2)}, \hat{y}'^{(2)})^T \cdot \nabla_{\theta_{2}}\ell(PL^{(1)}, \hat{y}^{(2)})$
\STATE $h^{(2)} \gets \nabla_{\theta'_{1}}\ell(Y_t^{(1)}, \hat{y}'^{(1)})^T \cdot \nabla_{\theta_{1}}\ell(PL^{(2)}, \hat{y}^{(1)})$
\STATE // Weights to be used in the next step
\STATE $\theta_{1} \gets \theta'_{1} - \eta \cdot h^{(1)} \cdot \nabla_{\theta_{1}}\ell(PL^{(1)}, \hat{y}^{(1)})$
\STATE $\theta_{2} \gets \theta'_{2} - \eta \cdot h^{(2)} \cdot \nabla_{\theta_{2}}\ell(PL^{(2)}, \hat{y}^{(2)})$
\ENDFOR 
\end{algorithmic}
\end{algorithm}

%
\ourparagraph{Algorithm Description}
Algorithm~\ref{alg:MCT} presents pseudocode for the algorithm.  The function \textit{SampleBatch}, is sampling a batch from the dataset. The function \textit{getOtherView} returns the complementary view of the first argument; the second argument clarifies which view should be returned (``2'' in line 5).
At each step $t$ of the algorithm each model is first updated based on the pseudo-labels the other has provided on a batch of unlabeled data (lines 13-14).  Then each model is updated to provide labels that encourage the other to predict more correctly on the labeled set based on the performance of the other (lines 24-25).  Further discussion of 
these update rules can be found in Appendix \ref{sec:background:MPL}.

Each student model evaluates the performance of their new weights on separate labeled batches.  This performance provides feedback to the teacher model.
Geometrically, we can understand this as updating the parameters in the direction of increasing the teacher's confidence on pseudo-labels, assuming those pseudo-labels helped the student model perform better on the labeled set. 
Along these lines, the teacher parameters are updated
in the opposite direction if it hurt the student's performance on the labeled set.

In general MCT does not require any preexisting method of view construction to be applied. If a dataset naturally provides two views, then there is no need to construct additional views. One can use their favorite pre-trained representation or none at all and then apply MCT as described above. Additionally we can relax the assumption that there is correspondence between the labeled examples for each view. If it happens to be the case that it is easy to collect unlabeled examples from each view which correspond to the same label but existing labeled sets for each view do not, then MCT can still be applied. This is because the labeled data used to evaluate the change in the performance of the student does not need to have corresponding instances in the teacher’s view.  In all of our experiments in this work we use preexisting foundation models to construct views for our co-training methods, and thus in all cases there exist two views for each labeled example.


In Algorithm~\ref{alg:MCT} it is illustrative to consider all versions of the parameters within a step and all of their gradients to exist simultaneously in memory. This is undesirable in practice given the large size of common neural networks and their gradients.  Instead, we compute $\nabla_{\theta_{2}}\ell(PL^{(2)}, \hat{y}^{(2)})$ and $\nabla_{\theta_{1}}\ell(PL^{(1)}, \hat{y}^{(1)})$ (used in lines 24-25) first, then we compute (and apply) $\nabla_{\theta_{1}}\ell(PL^{(2)}, \hat{y}^{(1)})$ and $\nabla_{\theta_{2}}\ell(PL^{(1)}, \hat{y}^{(2)})$ (used first in lines 13-14 and reused in lines 21-22), then we compute $\nabla_{\theta'_{1}}\ell(Y_t^{(1)}, \hat{y}'^{(1)})^T$ and $\nabla_{\theta'_{2}}\ell(Y_t^{(2)}, \hat{y}'^{(2)})^T$ which we subsequently use to compute $h^{(1)}$ and $h^{(2)}$. We finish the iteration with lines 24-25 using the saved gradient from the first step.  With this approach we only have to compute the forward pass and gradient for the student update once, and we only keep one version of the model weights in memory at a time. 

\ourparagraph{Relation to Ensemble Methods}

As our approach utilizes multiple models which collaborate during prediction, we acknowledge that it bears similarity to the family of ensemble methods. Typically, co-trained models are evaluated against other SSL approaches. In our evaluation we compare our semi-supervised method to supervised deep ensembles to illustrate the impact of MCT versus ensemble methods.  We show that when MCT works well, the performance benefit cannot be attributed to simply utilizing multiple models.

\section{Experimental Evaluation}
\label{sec:experiments}
Towards evaluating our proposed method we compare to other self-supervised and semi-supervised learning algorithms using the ImageNet~\cite{Dataset:ImageNet} classification benchmark, as well as Flowers102~\cite{Dataset:Flowers102}, Food101~\cite{Dataset:Food101}, FGVCAircraft~\cite{Dataset:FGVCAircraft}, iNaturalist~\cite{Dataset:iNaturalist}, and iNaturalist 2021~\cite{Dataset:iNat2021} datasets.
To produce the views used to train the classifiers during CT and MCT we used the embedding space of five representation learning architectures: The Masked Autoencoder (MAE) \cite{SSL:ST-MAE}, DINOv2 \cite{SSL:ST-DINOv2}, SwAV \cite{SSL:ST-SwAV}, EsViT \cite{SSL:ST-EsViT}, and CLIP~\cite{SSL:ST-CLIP}. We selected the models which produce the views as they have been learned in an unsupervised way, have been made available by the authors of their respective papers for use in PyTorch, and have been shown to produce representations that are appropriate for computer vision classification tasks.  Hyperparameters used during training are included in Table~\ref{tbl:hparam-config} in the appendix for CT, MCT, and deep ensembles. 

Table~\ref{tbl:data-sizes} 
provides a description of the different datasets used for training and testing, as well as the number of classes for each dataset. 
Note that in the 10\% experiments on the Flowers102 dataset each class has only 1 label.  
All datasets are approximately class-balanced with the exception of the iNaturalist dataset.  
In all cases we maintain the original class distribution when sampling subsets.  
For ImageNet we use the split published with the SimCLRv2~\cite{SSL:ST-SimCLRv2} repository.
All subsets are created with seeded randomness of a common seed (13) which ensures a fair comparison between methods.
\begin{table}[ht]
\caption{Characteristics of datasets used.}\label{tbl:data-sizes}
    \centering
    \begin{tabular}{lrrr}
        \toprule
        Dataset & \#train & \#test & \#classes \\
        \midrule
        FGVCAircraft & 3333 & 3333 & 100\\
        Flowers102 & 1020 & 6149  & 102\\
        Food101 & 75,750 & 25,250 & 101\\
        iNaturalist & 175,489 & 29,083 & 1010\\
        ImageNet & 1,281,167 & 50,000 & 1000\\
        \bottomrule
    \end{tabular}
\end{table}

\subsection{Experimental Evaluation on ImageNet}
\label{sec:CT-assumptions}

As we discussed in Section~\ref{sec:view-construction}, CT relies on two assumptions about the sufficiency and independence of the two views in our data. We conducted experiments to verify these assumptions.  Below we give a summary of our results.

\ourparagraph{On the Sufficiency of the Views}
Sufficiency is fairly easy to verify.  If we choose a reasonable sufficiency threshold for the task, for ImageNet say close or above $75\%$ top-1 accuracy, then we can train simple models on the views of the dataset we have constructed and provide examples of functions which demonstrate the satisfaction of the property.  We tested a single linear layer with softmax output 
to provide a lower bound on the sufficiency of the views on the standard subsets of the ImageNet labels for semi-supervised 
classification.


A simple 3-layer multi-layer perceptron (MLP) with 1024 neurons per layer was evaluated (Table~\ref{tab:MLP_exp}).
These were the same models as those used for CT and MCT.  From these experiments, whose results are shown in Table~\ref{tab:MLP_exp} we can see that if we were to pick a threshold for top-1 accuracy around $75\%$, 
then CLIP and DINOv2 provide sufficient views for both the $1\%$ and $10\%$ subsets.

\begin{table}[t]
\caption{MLP performance on individual views.  Top-1 accuracy on different subsets of the ImageNet data are shown.\hspace{\textwidth}}
    \label{tab:MLP_exp}
    \centering
    \begin{tabular}{lcc}
    \toprule
    Model & 1\% & 10\%\\
    \midrule
    MAE & 23.4 & 48.5 \\
    DINOv2 & 78.4 & 82.7 \\
    SwAV & 12.5 & 32.6 \\
    EsViT & 71.3 & 75.8\\
    CLIP & 75.2 & 80.9\\
    \bottomrule
    \end{tabular}
\end{table}

\ourparagraph{On the Independence of the Views}
To test the independence of the views generated by the representation learning models, we trained an identical MLP architecture to predict each view from each other view; Table~\ref{tab:pairwise_translation_ex} presents our findings, 
explained below.
\begin{table}[t]
\caption{Pairwise translation performance of linear probe on the ImageNet dataset.  A linear classifier is trained on the output of an MLP which is trained to predict one view (columns) from another view (rows) by minimizing MSE.  The top-1 accuracy (\%) of the linear classifier is reported.  Both the MLP and the linear classifier have access to the entire embedded ImageNet training set.}
    \label{tab:pairwise_translation_ex}
    \centering
    \resizebox{\columnwidth}{!}{
    \begin{tabular}{cccccc}
    \toprule
     & MAE & DINOv2 & SwAV & EsViT & CLIP\\
    \midrule
    MAE & - & 0.139 & 0.142 & 0.137 & 0.129 \\
    DINOv2 & 0.110 & - & 0.132 & 0.135 & 0.130\\
    SwAV & 0.116 & 0.147 & - & 0.137 & 0.126\\
    EsViT & 0.112 & 0.140 & 0.116 & - & 0.135 \\
    CLIP & 0.110 & 0.146 & 0.136 & 0.156 & - \\
    \bottomrule
    \end{tabular}
    }
\end{table}
Given as input the view to predict from (rows of Table~\ref{tab:pairwise_translation_ex}) the MLPs were trained to reduce the mean squared error (MSE) between their output and the view to be predicted (columns of Table~\ref{tab:pairwise_translation_ex}).  
We then trained a linear classifier on the outputs generated by the MLP given every training set embedding from 
each view.  
Had the model 
faithfully reconstructed the output view given only the input view then we would expect the linear classifier to perform 
similarly to 
the last column of Table~\ref{tab:linear_probe_exp}.  The linear classifiers never did much better than a random guess on the ImageNet class achieving at most 0.156\% accuracy.  
Thus, while we 
cannot immediately conclude that the views are independent, 
it is clearly not trivial to predict one view given any other.  We believe that this is compelling evidence for the independence of different representations.

\begin{table}[t]
    \caption{Linear probe evaluation for views. Top-1 accuracy on different subsets of the ImageNet data are shown.}\label{tab:linear_probe_exp}
    \centering
    \begin{tabular}{lccc}
    \toprule
    Model & 1\% & 10\% & 100\%\\
    \midrule
    MAE & 1.9 & 3.2 & 73.5\\
    DINOv2 & 78.1 & 82.9 & 86.3\\
    SwAV & 12.1 & 41.1 & 77.9\\
    EsViT & 69.1 & 74.4 & 81.3\\
    CLIP & 74.1 & 80.9 & 85.4\\
    \bottomrule
    \end{tabular}
\end{table}

Having constructed at least two strong views of the data, and suspecting that these views are independent we hypothesize that MCT and CT will work on this data.  


\ourparagraph{Co-Training Baseline Experiments}
At each iteration of CT the models made predictions for all instances in $U$.  We assigned labels to the 10\% of the most confident predictions in $U$ for both models.  When confident predictions conflicted on examples, we returned those instances to the unlabeled set, otherwise their complementary views were entered into the labeled set of the other model with the assigned pseudo-label.  

\begin{figure}[t]
    \centering
    \includegraphics[width=.45\textwidth]{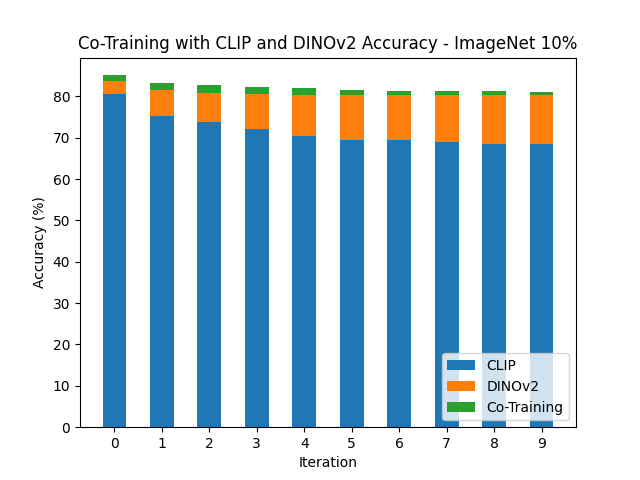}
    \caption{Top-1 accuracy of CT iterations on the CLIP and DINOv2 views for the ImageNet 10\% dataset.}
    \label{fig:CD-10}
    \bigskip
    \smallskip
\end{figure}

\begin{figure}[t]
    \centering
    \includegraphics[width=0.4\textwidth]{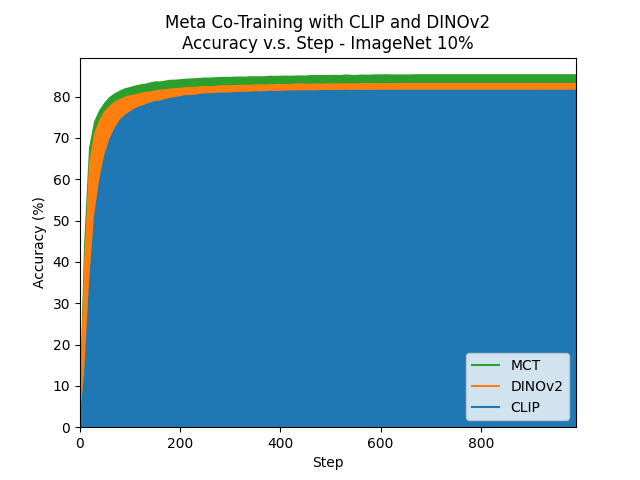}
    \caption{MCT using the CLIP and DINOv2 views as a function of the training step.  Models are trained on 10\% of the ImageNet labels.\\}\smallskip
    \label{fig:MCTCD10}
\end{figure}

For CT, joint predictions yielded better accuracy than the predictions of each individual model.  In our experiments on ImageNet, as CT iterations proceeded, top-1 accuracy decreased.  Later we show that this was not always the case (see Figure~\ref{fig:CTFood10}).  
The decrease was less pronounced when the views perform at a similar level; see Figure~\ref{fig:CD-10}. 
In contrast MCT does not exhibit performance degradation after warmup.
Figures~\ref{fig:CD-10} and~\ref{fig:MCTCD10} show results on ImageNet 10\%.  We see that while the CT performance decreases after the supervised warmup, MCT performance increases.  

Despite the poor performance of the pseudo-labeling during CT, 
using the two best performing views CT
\emph{performed as well as the previous SoTA method} for SSL classification on the ImageNet-10\%; see Co-Training in Table~\ref{tab:comparisons_exp}.  
CT as a method was rather disappointing in that the algorithm is unable to leverage pseudo-labels to improve the accuracy on this task.
Given the difficulty of translating between the views, and the fact that the accuracy yielded by the joint prediction 
is 
greater than its constituent views, we believe the poor performance of the pseudo-labeling step in CT was due to 
imbalanced information content between views rather than view-dependence. 

\begin{table*}[ht]
\caption{Performance of different approaches on ImageNet dataset. An asterisk (\textcolor{black}{*}) indicates models that were trained on top of embeddings generated by much larger models.  During training we do not need to alter the parameters of these larger models, and these larger models need not see data used for downstream classification during their training, however training of downstream models would not have been possible without the embeddings.  Models on the lower half of the table use unlabeled data during classification training.}
\label{tab:comparisons_exp}
    \centering
    \begin{tabular}{cllcc}
        \toprule
         Reference & \multicolumn{1}{l}{Model} & Method & ImageNet-1\% & ImageNet-10\%\\
        \midrule
        \cite{SSL:ST-EsViT} & EsViT (Swin-B, W=14) & Linear & 69.1 & 74.4\\
        \cite{SSL:ST-MAE} & MAE (ViT-L) & MLP & 23.4 & 48.5 \\
        \cite{SSL:ST-CLIP} & CLIP (ViT-L) & Fine-tuned & 80.5 & 84.7\\
        \cite{SSL:ST-DINOv2} & DINOv2 (ViT-L) & Linear & 78.1 & 82.9 \\
        \cite{SSL:ST-SimCLRv2} & SimCLRv2 (ResNet152-w2) & Fine-tuned & 74.2 & 79.4\\
        \cite{SSL:ST-SwAV} & SwAV (RN50-w4) & Fine-tuned & 53.9 & 70.2\\
        \cmidrule(lr{.75em}){1-5}
        \cite{SSL:co-training:deep}& Deep Co-Training (ResNet-18) & Co-Training & - & 53.5 \\
        \cite{SSL:UDA} & UDA (ResNet50) & UDA & - & 68.78\\
        \cite{SSL:FixMatch}& FixMatch (ResNet-50) & FixMatch& - & 71.46\\
        \cite{SSL:MPL}& MPL (EfficientNet-B6-Wide) & MPL & - & 73.9 \\
        \cite{SSL:ST-Semi}& Semi-ViT (ViT-L) & Self-trained & 77.3 & 83.3 \\
        \cite{REACT}& REACT (ViT-L) & REACT & \textbf{81.6} & 85.1 \\
        \cite{SSL:co-training}& Co-Training (MLP)\textcolor{black}{*} & Co-Training & 80.1 & 85.1
        \\
        \cite{SL:DeepEnsembles}& Deep Ensemble \textcolor{black}{*} & Deep Ensemble & 80.0 & 84.3 \\
        & Meta Co-Training (MLP)\textcolor{black}{*} - ours & Meta Co-Training & 80.7 
        & \textbf{85.8} 
        \\
        \bottomrule
    \end{tabular}
\end{table*}

\ourparagraph{Meta Co-Training Experiments}
To draw fair comparisons, we fix the model architecture and view set from our experiments on CT.  
As in MPL~\cite{SSL:MPL} a supervised loss is optimized jointly with a loss on pseudo-labels.
We found that beginning with a \emph{warmup} period in which the models for each view are trained in a strictly supervised way expedited training.  This warmup period occurs for the same number of updates as the first training phase of CT, so each method starts with the same number of supervised updates before pseudo-labeling.  
Unlike MPL, we did not use Unsupervised Data Augmentation~\cite{SSL:UDA}, because we embedded the dataset only once, and using self-supervised models to generate the views we did not believe data augmentation would have much of an effect.  
Details of our training recipe are given in Table~\ref{tbl:hparam-config}.

Table \ref{tbl:CT_MCT_exp} shows the results of MCT on all of the possible combinations of views we used.  With one exception, the better the performance of the views, the better MCT performs compared to CT.  In nearly all cases in which both views are strong, MCT performs better than CT.  These are also the cases with the highest performance overall.  Comparing CT accuracy on ImageNet 10\% (resp.~1\%) shown in Table~\ref{tab:CoTrain_exp} compared to MLP accuracy shown in Table~\ref{tab:MLP_exp}, we observe that CT top-1 accuracy is on average 18.7\% (resp.~16.8\%) higher than MLP accuracy. Moreover, the best CT top-1 accuracy is 2.4\% (resp.~1.7\%) higher
compared to the best top-1 MLP accuracy.  In Table~\ref{tab:DE_exp} we show the performance of an ensemble of models trained on concatenated view pairs but we provide more information on the comparison between MCT and deep ensembles in a separate paragraph below.

\begin{table*}[ht]
\caption{CT, MCT, and ensemble top-1 accuracy of view combinations on 10\% (1\%) of ImageNet labels.  As CT and MCT do not depend on the order of the views, we show only upper-diagonal entries of the pairwise comparison.  
All $\binom{5}{2}=10$ different combinations of views are shown in each table. In the case of the MLP ensemble the order of the pairs of views also does not matter.  These views are concatenated in order to form a larger unified view and hence different order of the views would correspond to different order of the attributes in the embeddings, which does not change their information content.\\}\label{tbl:CT_MCT_exp}
\begin{subtable}[t]{0.49\textwidth}
    \caption{Co-Training}
    \label{tab:CoTrain_exp}
    \centering
    \resizebox{\columnwidth}{!}{
    \begin{tabular}{ccccc}
    \toprule
     & DINOv2 & SwAV & EsViT & CLIP\\
    \midrule
    MAE      & {81.8} ({75.5}) & {30.5} ({11.4}) & {77.1} ({66.2}) & {78.8} ({66.8}) \\
    DINOv2   &             & {78.5} ({74.5}) & {83.3} ({78.9})  & \textbf{{85.1}} (\textbf{{80.1}}) \\
    SwAV     &             &             & {70.6} ( {64.9})  &  {75.5} (\textcolor{black}{67.4}) \\
    EsViT    &             &             &              &  {82.3} ( {76.9}) \\
    \bottomrule
    \end{tabular}
    }
\end{subtable}
\hspace{\fill}
\begin{subtable}[t]{0.49\textwidth}
\caption{Meta Co-Training}\label{tab:MCT_exp}
\resizebox{\columnwidth}{!}{
\begin{tabular}{ccccc}
    \toprule
     & DINOv2 & SwAV & EsViT & CLIP\\
    \midrule
    MAE      &  {81.2} ({73.8}) & {37.6} ({8.1})  &  {73.6} ( {63.8}) & {78.6} ({63.5}) \\
    DINOv2   &             &  {77.3} ( {70.7}) & {83.4} ({79.2}) & \textbf{{85.2}} (\textbf{{80.7}}) \\
    SwAV     &             &             & {74.4} ({67.3}) & {77.1} (67.4) \\
    EsViT    &             &             &             & {82.4} ({77.5}) \\
    \bottomrule
\end{tabular}
}
\end{subtable}
\vspace{8pt}
\begin{center}
\begin{subtable}[t]{0.49\textwidth}
\caption{MLP Ensemble}\label{tab:DE_exp}
\centering
\resizebox{\columnwidth}{!}{
\begin{tabular}{ccccc}
    \toprule
     & DINOv2 & SwAV & EsViT & CLIP\\
    \midrule
    MAE      &  {83.0} ({78.9}) & {38.1} ({17.9})  &  {75.0} ( {72.1}) & {79.4} ({76.1}) \\
    DINOv2   &             &  {83.7} ( {79.1}) & {81.2} ({78.5}) & \textbf{{84.2}} (\textbf{{80.0}}) \\
    SwAV     &             &             & 74.4 ({72.5}) & {81.2} ({76.8}) \\
    EsViT    &             &             &             & {78.7} ({73.5}) \\
    \bottomrule
\end{tabular}
}
\end{subtable}
\end{center}
\end{table*}

MCT and CT are compared on ImageNet 10\% (resp.,~1\%) in 
Table~\ref{tab:comparisons_exp}.
We observe that MCT top-1 accuracy is 0.7\% (resp.,~0.6\%) higher overall.  An ablation study for the views is given in Table~\ref{tbl:CT_MCT_exp}.
On all view pairs in which MCT or CT provide no benefit over a single view are 
cases in which one of the views was significantly weaker than the other.  This makes sense as one model has nothing to learn from the other.  For the experiments in which we had two strong views, MCT always outperforms CT.  

\ourparagraph{Comparison to Ensemble Methods}
Ensemble methods are common in practice to boost the performance of supervised models.  To show that our method provides benefit outside of just adding an additional model, we compare to a Deep Ensemble \cite{SL:DeepEnsembles} of five models for each individual view.  In Table~\ref{tab:MLP_ens_exp} we show the performance of ensembles trained on the individual views. We show the performance of an ensemble trained on each view pair concatenated in Table~\ref{tab:DE_exp}. We show the performance of an ensemble trained on the concatenation of all four views we used to achieve our best performing MCT experiment in Table~\ref{tab:CTMCT_exp}.  

In some cases the CT models perform better than the ensemble of models which all have the same input for each sample (the concatenation of the CT views).  We believe that CT before pseudo-labeling (which is essentially an ensemble of two different views) out-performs the deep ensembles because the CT models are fewer but they are more diverse.
The CT models are more diverse as they are trained on different views, while the ensemble models are trained on the \emph{concatenation} of the views. The concatenation has the same information but may not produce meaningfully different models in an ensemble.
Ensuring diversity of ensemble members is an important problem for ensembling methods, so it seems likely that a less diverse ensemble would perform worse.  If this is indeed the case, then it would support our hypothesis that CT is failing to improve after pseudo-labeling due to an imbalance in the sufficiency of views and not a lack of independence.  MCT provides performance gains over CT and the deep ensembles, so we can be confident that these gains cannot be attributed solely to the use of multiple models.

\begin{table}[t]
\caption{MLP ensemble performance on individual views.  Top-1 accuracy on different subsets of the ImageNet data are shown.\hspace{\textwidth}}
    \label{tab:MLP_ens_exp}
    \centering
    \begin{tabular}{lcc}
    \toprule
    Model & 1\% & 10\%\\
    \midrule
    MAE & 1.7 & 3.8 \\
    DINOv2 & 79.2 & 82.8 \\
    SwAV & 19.7 & 40.3 \\
    EsViT & 72.1 & 75.0\\
    CLIP & 76.4 & 80.8\\
    \bottomrule
    \end{tabular}
\end{table}

\subsection{Stronger Views by Concatenation}
Finally, we take the four views which had the greatest performance and constructed two views out of them by concatenating them together.  We constructed one view 
CLIP $\vert$ EsViT
and the second 
DINOv2 $\vert$ SwAV
and measured the performance of CT and MCT on these views in Table~\ref{tab:CTMCT_exp}.  Interestingly, CT does not benefit from these larger views; see Table~\ref{tab:CTMCT_exp}.  
\begin{table}[t]
\caption{Co-training and MCT evaluated on the ImageNet dataset using the CLIP $\vert$ EsViT and DINOv2 $\vert$ SwAV views. The Deep Ensemble is evaluated on the concatenation of all four views.}
\label{tab:CTMCT_exp}
    \centering
    \begin{tabular}{ccc}
    \toprule
    Method & 1\% & 10\%\\
    \midrule
    Deep Ensemble & 80.0 & 84.3 \\
    Co-Training & 80.0 & 84.8 \\
    Meta Co-Training & \textbf{80.5} & \textbf{85.8} \\
    \bottomrule
    \end{tabular}
\end{table}

The additional information may have allowed the individual models to overfit their training data quickly.  During MCT we observed that after one model reaches 100\% training accuracy, its validation accuracy can still improve by learning to label for the other model. Possibly due to the rapid overfitting due to the increased view size (and consequently increased parameter count) the 1\% split did not show improvement for either model.

\subsection{Experiments on Additional Datasets}\label{sec:main:more-datasets}
\begin{figure}[ht]
    \centering
    \includegraphics[width=0.45\textwidth]{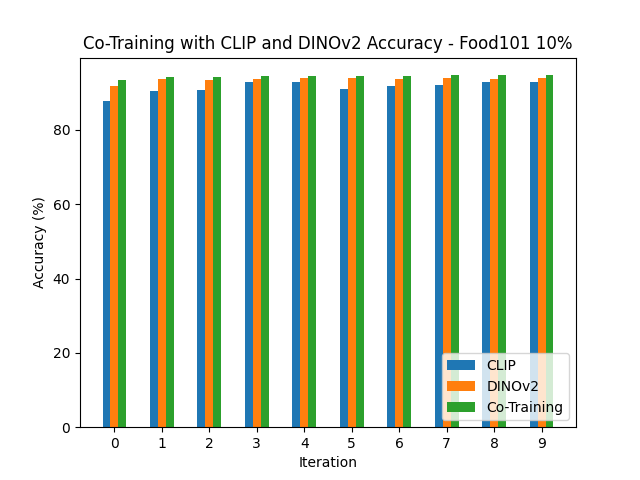}
    \caption{The top-1 accuracy of the CT predictions for each iteration of CT.  10\% of available Food101 labels are used for training.  The method exhibits performance improvement over multiple iterations of pseudo-labeling and retraining.}
    \label{fig:CTFood10}
\end{figure}

 We compare MCT and CT to existing SoTA approaches that leverage unlabeled data directly during training (e.g.,~not just part of a pretext task).  These experiments support our hypothesis that the primary cause of the failure of CT is 
that one or more of the views is sufficient to learn a model that achieves high accuracy. 
The closer to equivalent in performance individual views are, the less performance suffers. When the two views perform similarly and well, the performance improves.  
%
Tables~\ref{tab:REACT+} and~\ref{tab:Semi+}
provide more information on these additional datasets. 
Furthermore, Figure~\ref{fig:CTFood10} provides an example where CT performs as expected and accuracy increases across CT iterations.
In all but one case MCT outperforms the SoTA top-1 accuracy.  The main takeaway of these additional experiments, apart from establishing new SoTA results in datasets beyond ImageNet, is that 
MCT is in general more robust to view imbalance and provides better results than CT.
As the performance degradation of co-training beyond the initial warmup period was not something expected, in 
Appendix~\ref{sec:appendix:CT} we provide charts similar to the one in Figure~\ref{fig:CTFood10}, but for the other cases of datasets or presence of labels in the training set, thus, giving full details on the results that we obtained for CT.

\begin{table}[t]
\caption{Additional comparisons to REACT.  The authors include experiments for zero-shot performance and 10\% of available labels.  Results shown are for 10\% of labels.  For Flowers102 this is 1-shot performance.}
    \label{tab:REACT+}
    \centering
    \resizebox{\columnwidth}{!}{
    \begin{tabular}{lccc}
    \toprule
    & \multicolumn{3}{c}{Dataset} \\
    \cmidrule(lr{.75em}){2-4}
    Method & Flowers102 & Food101 & FGVCAircraft \\
    \midrule
    REACT (ViT-L) & 97.0 & 85.6 & \textbf{57.1}\\
    Co-Training (MLP) & 99.2 & 94.7 & 36.4\\
    Meta Co-Training (MLP) & \textbf{99.6} & \textbf{94.8} & 40.1\\
    \bottomrule
    \end{tabular}
    }
\end{table}
\begin{table}[t]
\caption{Additional comparisons to Semi-ViT using 10\% (1\%) of the available labels for training.  For the iNaturalist task we use only the 1010 most frequent classes.}
    \label{tab:Semi+}
    \centering
    \begin{tabular}{lcc}
    \toprule
    & \multicolumn{2}{c}{Dataset} \\
    \cmidrule(lr{.75em}){2-3}
    Method & iNaturalist & Food101 \\
    \midrule
    Semi-ViT (ViT-B) & 67.7 (32.3) & 84.5 (60.9)\\
    Co-Training (MLP) & 59.7 (29.5) & 94.7 (83.9)\\
    Meta Co-Training (MLP) & \textbf{76.0} (\textbf{58.1}) & \textbf{94.8} (\textbf{91.7})\\
    \bottomrule
    \end{tabular}
\end{table}

\section{Related Work}
\label{sec:related}

While SSL
has become very relevant in recent years with the influx of the data 
outpacing the availability of human labels, the paradigm and its usefulness have been studied at least since the 1960's; e.g.,~\cite{SSL:old1,SSL:old2,SSL:ST-old1,SSL:ST-old2}.
%
Among the main approaches for SSL that are not directly related to our work, one popular direction encourages entropy minimization and consistency regularization using data augmentation such as MixMatch~\cite{SSL:MixMatch}, UDA~\cite{SSL:UDA}, ReMixMatch~\cite{SSL:ReMixMatch}, FixMatch~\cite{SSL:FixMatch}, and FlexMatch~\cite{SSL:FlexMatch}.

\ourparagraph{Self-Training and Student-Teacher Frameworks}
One of the earliest ideas for SSL is that of \emph{self-training}~\cite{SSL:ST-old1,SSL:ST-old2}. This idea is still very popular with different \emph{student-teacher} frameworks~\cite{SSL:Rethinking,SSL:NST,SSL:MPL,SSL:Meta:LST,SSL:ST-hierarchical,SSL:ST-Semi}. 
This technique is also frequently referred to as \emph{pseudo-labeling}~\cite{SSL:ST:PL} as the student tries to improve its performance by making predictions on the unlabeled data, then integrating the most confident predictions to the labeled set, and subsequently retraining using the broader labeled set.
In addition, one can combine ideas; e.g., self-training with clustering~\cite{SSL:ST-Clustering:Mean-Shift}, or create ensembles via self-training~\cite{SSL:ST-Self-Ensemble}, potentially using graph-based information~\cite{SSL:ST-SNTG}.
An issue with these approaches is that of  \emph{confirmation bias}~\cite{SSL:ConfirmationBias}, where the student cannot improve a lot due to incorrect labeling that occurs either because of self-training, or due to a fallible teacher.
\ourparagraph{Co-training Algorithms}
Co-training was introduced by Blum and Mitchell~\cite{SSL:co-training} and their paradigmatic example was classifying web pages using two different views: \emph{(i)} the 
\emph{bags of words} based on the content that the webpages had, and \emph{(ii)}
the bag of words formed by the words 
used in hyperlinks pointing to these webpages. In our context the two different views are obtained via different \emph{embeddings} that we obtain when we use images as inputs to different foundation models and observe the resulting embeddings.

Co-training has been very successful with wide applicability~\cite{SSL:co-training:verifying} and can outperform learning algorithms that use only single-view data~\cite{SSL:co-training:effectiveness}. Furthermore, the generalization ability of co-training is well-motivated theoretically~\cite{SSL:co-training,SSL:co-training-complexity,SSL:co-training-complexityBalcan,SSL:co-training-complexityBalazs}.
There are several natural extensions of the base idea, such as methods for constructing different views~\cite{SSL:co-training:view-construction:RSM,SSL:co-training:view-construction:RandomSubset,SSL:co-training:view-construction:AutoPart,SSL:co-training:view-construction:RandomSubsetEM}, 
including connections to active learning~\cite{SSL:co-training:view-construction:ActiveDisagreement}, methods that are specific to deep learning~\cite{SSL:co-training:deep}, and other 
extensions that exploit more than two views on data; e.g.,~\cite{SSL:co-training:multiple-views:tri-training}. 

\section{Conclusion and Future Work}
\label{sec:conclusion}
We proposed MCT, a novel semi-supervised learning algorithm. 
We showed that our algorithm outperforms co-training in general and establishes a new SoTA top-1 classification accuracy on the ImageNet 10\%. 
We also tested the performance of deep ensembles on ImageNet 10\% and 1\%. In both cases MCT achieved better results, demonstrating that MCT provides utility beyond what can be attributed to ensembling.
In addition, using MCT, we were able to achieve 
new SoTA few-shot top-1 classification accuracy on 
other 
popular CV benchmark datasets, including one-shot top-1 classification accuracy for the Flowers102 dataset.
Furthermore, we presented evidence that self-supervised learned representations are good candidates for views.  We showed that the views constructed this way 
likely contain independent information and that models trained on these views, or concatenation of these views, can complement each other to produce stronger classifiers.  

Our experiments indicate that our method performs best when it has access to enough labeled data to produce strong views for pseudo-labeling.  For some datasets this may preclude its application in one-shot settings.  Further, if no sufficient and independent views are available and none can be constructed then we do not expect our model to perform meaningfully better than MPL.

For future work we expect that fine-tuning the representation learners that we used for CT and 
MCT 
would improve results. Papers proposing foundation models observe performance benefits when fine-tuning those models on downstream tasks.
Furthermore, we did not make use of the large available unlabeled datasets for CV in this study.  
Giving MCT access to a large repository of unlabeled data, as well as the entirety of the labeled data, should result in a performance boost.  Finally, our list of pretext tasks was not exhaustive.  It is an interesting question for future work which other tasks and models might yield useful views.

\section{Acknowledgments}
This material is based upon work supported by the U.S.~National Science Foundation under Grant No.~RISE-2019758.
This work is part of the NSF AI Institute for Research on Trustworthy AI in Weather, Climate, and Coastal Oceanography (NSF AI2ES).

Some of the computing for this project was performed at the OU Supercomputing Center for Education \& Research (OSCER) at the University of Oklahoma (OU).

{\small
\bibliographystyle{ieee_fullname}
\bibliography{references}

\begin{thebibliography}{10}\itemsep=-1pt

\bibitem{SSL:ConfirmationBias}
Eric Arazo, Diego Ortego, Paul Albert, Noel~E. O'Connor, and Kevin McGuinness.
\newblock {Pseudo-Labeling and Confirmation Bias in Deep Semi-Supervised Learning}.
\newblock In {\em IJCNN}, pages 1--8. {IEEE}, 2020.

\bibitem{SSL:co-training-complexityBalcan}
Maria{-}Florina Balcan, Avrim Blum, and Ke Yang.
\newblock {Co-Training and Expansion: Towards Bridging Theory and Practice}.
\newblock In {\em NeurIPS}, pages 89--96, 2004.

\bibitem{SSL:ReMixMatch}
David Berthelot, Nicholas Carlini, Ekin~D. Cubuk, Alex Kurakin, Kihyuk Sohn, Han Zhang, and Colin Raffel.
\newblock {ReMixMatch: Semi-Supervised Learning with Distribution Alignment and Augmentation Anchoring}.
\newblock {\em CoRR}, abs/1911.09785, 2019.

\bibitem{SSL:MixMatch}
David Berthelot, Nicholas Carlini, Ian~J. Goodfellow, Nicolas Papernot, Avital Oliver, and Colin Raffel.
\newblock {MixMatch: {A} Holistic Approach to Semi-Supervised Learning}.
\newblock In {\em NeurIPS}, pages 5050--5060, 2019.

\bibitem{SSL:co-training}
Avrim Blum and Tom~M. Mitchell.
\newblock {Combining Labeled and Unlabeled Data with Co-Training}.
\newblock In {\em COLT}, pages 92--100. {ACM}, 1998.

\bibitem{Dataset:Food101}
Lukas Bossard, Matthieu Guillaumin, and Luc~Van Gool.
\newblock {Food-101 - Mining Discriminative Components with Random Forests}.
\newblock In {\em ECCV}, pages 446--461, 2014.

\bibitem{SSL:co-training:view-construction:RandomSubset}
Ulf Brefeld, Christoph B{\"{u}}scher, and Tobias Scheffer.
\newblock {Multi-view Discriminative Sequential Learning}.
\newblock In {\em ECML}, pages 60--71. Springer, 2005.

\bibitem{SSL:co-training:view-construction:RandomSubsetEM}
Ulf Brefeld and Tobias Scheffer.
\newblock {Co-EM support vector learning}.
\newblock In {\em ICML}. {ACM}, 2004.

\bibitem{SSL:ST-Semi}
Zhaowei Cai and et al.
\newblock {Semi-supervised Vision Transformers at Scale}.
\newblock In {\em NeurIPS}, 2022.

\bibitem{SSL:ST-SwAV}
Mathilde Caron, Ishan Misra, Julien Mairal, Priya Goyal, Piotr Bojanowski, and Armand Joulin.
\newblock {Unsupervised Learning of Visual Features by Contrasting Cluster Assignments}.
\newblock In {\em NeurIPS}, 2020.

\bibitem{SSL:co-training:view-construction:AutoPart}
Minmin Chen, Kilian~Q. Weinberger, and Yixin Chen.
\newblock {Automatic Feature Decomposition for Single View Co-training}.
\newblock In {\em ICML}, pages 953--960. Omnipress, 2011.

\bibitem{SSL:ST-SimCLR}
Ting Chen, Simon Kornblith, Mohammad Norouzi, and Geoffrey~E. Hinton.
\newblock {A Simple Framework for Contrastive Learning of Visual Representations}.
\newblock In {\em ICML}, pages 1597--1607. {PMLR}, 2020.

\bibitem{SSL:ST-SimCLRv2}
Ting Chen, Simon Kornblith, Kevin Swersky, Mohammad Norouzi, and Geoffrey~E. Hinton.
\newblock {Big Self-Supervised Models are Strong Semi-Supervised Learners}.
\newblock In {\em NeurIPS}, 2020.

\bibitem{SSL:co-training-complexityBalazs}
Malte Darnst{\"{a}}dt, Hans~Ulrich Simon, and Bal{\'{a}}zs Sz{\"{o}}r{\'{e}}nyi.
\newblock {Supervised learning and Co-training}.
\newblock {\em Theor. Comput. Sci.}, 519:68--87, 2014.

\bibitem{SSL:co-training-complexity}
Sanjoy Dasgupta, Michael~L. Littman, and David~A. McAllester.
\newblock {PAC Generalization Bounds for Co-training}.
\newblock In {\em NeurIPS}, pages 375--382. {MIT} Press, 2001.

\bibitem{SSL:old2}
Neil~E Day.
\newblock {Estimating the components of a mixture of normal distributions}.
\newblock {\em Biometrika}, 56(3):463--474, 1969.

\bibitem{Dataset:ImageNet}
Jia Deng, Wei Dong, Richard Socher, Li{-}Jia Li, Kai Li, and Li Fei{-}Fei.
\newblock {ImageNet: {A} large-scale hierarchical image database}.
\newblock In {\em CVPR}, pages 248--255. {IEEE}, 2009.

\bibitem{SSL:ST-BERT}
Jacob Devlin, Ming{-}Wei Chang, Kenton Lee, and Kristina Toutanova.
\newblock {{BERT:} Pre-training of Deep Bidirectional Transformers for Language Understanding}.
\newblock In {\em NAACL-HLT}, pages 4171--4186. ACL, 2019.

\bibitem{SSL:co-training:view-construction:ActiveDisagreement}
Wei Di and Melba~M. Crawford.
\newblock {View Generation for Multiview Maximum Disagreement Based Active Learning for Hyperspectral Image Classification}.
\newblock {\em {IEEE} Trans. Geosci. Remote. Sens.}, 50(5-2):1942--1954, 2012.

\bibitem{SSL:co-training:verifying}
Jun Du, Charles~X. Ling, and Zhi-Hua Zhou.
\newblock {When Does Cotraining Work in Real Data?}
\newblock {\em IEEE Trans. on Knowledge and Data Engineering}, 23(5):788--799, 2011.

\bibitem{SSL:MPL:Approximation2}
Chelsea Finn, Pieter Abbeel, and Sergey Levine.
\newblock {Model-Agnostic Meta-Learning for Fast Adaptation of Deep Networks}.
\newblock In {\em ICML}, pages 1126--1135, 2017.

\bibitem{SSL:ST-old2}
S Fralick.
\newblock {Learning to recognize patterns without a teacher}.
\newblock {\em IEEE Trans. on Inf. Th.}, 13(1):57--64, 1967.

\bibitem{SSL:ST-BYOL}
Jean{-}Bastien Grill and et al.
\newblock {Bootstrap Your Own Latent - {A} New Approach to Self-Supervised Learning}.
\newblock In {\em NeurIPS}, 2020.

\bibitem{SSL:ST-hierarchical}
Xiaowei Gu.
\newblock {A self-training hierarchical prototype-based approach for semi-supervised classification}.
\newblock {\em Inf. Sci.}, 535:204--224, 2020.

\bibitem{SSL:old1}
Herman~Otto Hartley and Jon~NK Rao.
\newblock {Classification and estimation in analysis of variance problems}.
\newblock {\em Revue de l'Inst. Intl. de Stat.}, pages 141--147, 1968.

\bibitem{SSL:ST-MAE}
Kaiming He, Xinlei Chen, Saining Xie, Yanghao Li, Piotr Doll{\'{a}}r, and Ross~B. Girshick.
\newblock {Masked Autoencoders Are Scalable Vision Learners}.
\newblock In {\em CVPR}, pages 15979--15988. {IEEE}, 2022.

\bibitem{SSL:ST-MoCo}
Kaiming He, Haoqi Fan, Yuxin Wu, Saining Xie, and Ross~B. Girshick.
\newblock {Momentum Contrast for Unsupervised Visual Representation Learning}.
\newblock In {\em CVPR}, pages 9726--9735. Computer Vision Foundation / {IEEE}, 2020.

\bibitem{Dataset:iNat2021}
Grant~Van Horn, Elijah Cole, Sara Beery, Kimberly Wilber, Serge~J. Belongie, and Oisin~Mac Aodha.
\newblock {Benchmarking Representation Learning for Natural World Image Collections}.
\newblock In {\em CVPR}, pages 12884--12893, 2021.

\bibitem{Dataset:iNaturalist}
Grant~Van Horn and et al.
\newblock {The INaturalist Species Classification and Detection Dataset}.
\newblock In {\em CVPR}, pages 8769--8778, 2018.

\bibitem{SSL:ST-Clustering:Mean-Shift}
Soroush~Abbasi Koohpayegani, Ajinkya Tejankar, and Hamed Pirsiavash.
\newblock {Mean Shift for Self-Supervised Learning}.
\newblock In {\em ICCV}, pages 10306--10315. {IEEE}, 2021.

\bibitem{SSL:ST-Self-Ensemble}
Samuli Laine and Timo Aila.
\newblock {Temporal Ensembling for Semi-Supervised Learning}.
\newblock In {\em ICLR}, 2017.

\bibitem{SL:DeepEnsembles}
Balaji Lakshminarayanan, Alexander Pritzel, and Charles Blundell.
\newblock {Simple and Scalable Predictive Uncertainty Estimation using Deep Ensembles}.
\newblock In {\em NeurIPS}, pages 6402--6413, 2017.

\bibitem{SSL:ST:PL}
Dong-Hyun Lee.
\newblock {Pseudo-Label : The Simple and Efficient Semi-Supervised Learning Method for Deep Neural Networks}.
\newblock {\em Workshop on Challenges in Representation Learning, ICML}, 3(2):896, 2013.

\bibitem{SSL:ST-EsViT}
Chunyuan Li and et al.
\newblock {Efficient Self-supervised Vision Transformers for Representation Learning}.
\newblock In {\em ICLR}, 2022.

\bibitem{SSL:Meta:LST}
Xinzhe Li and et al.
\newblock {Learning to Self-Train for Semi-Supervised Few-Shot Classification}.
\newblock In {\em NeurIPS}, pages 10276--10286, 2019.

\bibitem{SSL:MPL:Approximation1}
Hanxiao Liu, Karen Simonyan, and Yiming Yang.
\newblock {{DARTS:} Differentiable Architecture Search}.
\newblock In {\em ICLR}. OpenReview.net, 2019.

\bibitem{REACT}
Haotian Liu, Kilho Son, Jianwei Yang, Ce Liu, Jianfeng Gao, Yong~Jae Lee, and Chunyuan Li.
\newblock {Learning Customized Visual Models with Retrieval-Augmented Knowledge}.
\newblock In {\em CVPR}, pages 15148--15158. {IEEE}, 2023.

\bibitem{SSL:ST-SNTG}
Yucen Luo, Jun Zhu, Mengxi Li, Yong Ren, and Bo Zhang.
\newblock {Smooth Neighbors on Teacher Graphs for Semi-Supervised Learning}.
\newblock In {\em CVPR}, pages 8896--8905, 2018.

\bibitem{Dataset:FGVCAircraft}
S. Maji, J. Kannala, E. Rahtu, M. Blaschko, and A. Vedaldi.
\newblock {Fine-Grained Visual Classification of Aircraft}.
\newblock Technical report, VGG, University of Oxford, 2013.

\bibitem{SSL:co-training:effectiveness}
Kamal Nigam and Rayid Ghani.
\newblock {Analyzing the Effectiveness and Applicability of Co-training}.
\newblock In {\em CIKM}, pages 86--93. {ACM}, 2000.

\bibitem{Dataset:Flowers102}
Maria{-}Elena Nilsback and Andrew Zisserman.
\newblock {Automated Flower Classification over a Large Number of Classes}.
\newblock In {\em ICVGIP}, pages 722--729, 2008.

\bibitem{SSL:ST-DINOv2}
Maxime Oquab and et al.
\newblock {DINOv2: Learning Robust Visual Features without Supervision}.
\newblock {\em Trans. Mach. Learn. Res.}, 2024, 2024.

\bibitem{SSL:MPL}
Hieu Pham, Zihang Dai, Qizhe Xie, and Quoc~V. Le.
\newblock {Meta Pseudo Labels}.
\newblock In {\em CVPR}, pages 11557--11568, 2021.

\bibitem{SSL:co-training:deep}
Siyuan Qiao, Wei Shen, Zhishuai Zhang, Bo Wang, and Alan~L. Yuille.
\newblock {Deep Co-Training for Semi-Supervised Image Recognition}.
\newblock In {\em ECCV}, pages 142--159, 2018.

\bibitem{SSL:ST-CLIP}
Alec Radford and et al.
\newblock {Learning Transferable Visual Models From Natural Language Supervision}.
\newblock In {\em ICML}, pages 8748--8763, 2021.

\bibitem{LAION400M}
Christoph Schuhmann and et al.
\newblock {{LAION-400M:} Open Dataset of CLIP-Filtered 400 Million Image-Text Pairs}.
\newblock In {\em Data Centric AI NeurIPS Workshop}, 2021.

\bibitem{LAION5B}
Christoph Schuhmann and et al.
\newblock {{LAION-5B:} An open large-scale dataset for training next generation image-text models}.
\newblock In {\em NeurIPS}, 2022.

\bibitem{SSL:ST-old1}
Henry Scudder.
\newblock {Probability of error of some adaptive pattern-recognition machines}.
\newblock {\em IEEE Trans. on Inf. Th.}, 11(3):363--371, 1965.

\bibitem{SSL:FixMatch}
Kihyuk Sohn, David Berthelot, Nicholas Carlini, Zizhao Zhang, Han Zhang, Colin Raffel, Ekin~Dogus Cubuk, Alexey Kurakin, and Chun{-}Liang Li.
\newblock {FixMatch: Simplifying Semi-Supervised Learning with Consistency and Confidence}.
\newblock In {\em NeurIPS}, 2020.

\bibitem{SSL:co-training:view-construction:RSM}
Jiao Wang, Siwei Luo, and Xianhua Zeng.
\newblock {A random subspace method for co-training}.
\newblock In {\em IJCNN}, pages 195--200. {IEEE}, 2008.

\bibitem{SSL:UDA}
Qizhe Xie, Zihang Dai, Eduard~H. Hovy, Thang Luong, and Quoc Le.
\newblock {Unsupervised Data Augmentation for Consistency Training}.
\newblock In {\em NeurIPS}, 2020.

\bibitem{SSL:NST}
Qizhe Xie, Minh{-}Thang Luong, Eduard~H. Hovy, and Quoc~V. Le.
\newblock {Self-Training With Noisy Student Improves ImageNet Classification}.
\newblock In {\em CVPR}, pages 10684--10695, 2020.

\bibitem{SSL:FlexMatch}
Bowen Zhang, Yidong Wang, Wenxin Hou, Hao Wu, Jindong Wang, Manabu Okumura, and Takahiro Shinozaki.
\newblock {FlexMatch: Boosting Semi-Supervised Learning with Curriculum Pseudo Labeling}.
\newblock In {\em NeurIPS}, pages 18408--18419, 2021.

\bibitem{SSL:co-training:multiple-views:tri-training}
Zhi{-}Hua Zhou and Ming Li.
\newblock {Tri-Training: Exploiting Unlabeled Data Using Three Classifiers}.
\newblock {\em {IEEE} Trans. Knowl. Data Eng.}, 17(11):1529--1541, 2005.

\bibitem{SSL:Rethinking}
Barret Zoph, Golnaz Ghiasi, Tsung{-}Yi Lin, Yin Cui, Hanxiao Liu, Ekin~Dogus Cubuk, and Quoc Le.
\newblock {Rethinking Pre-training and Self-training}.
\newblock In {\em NeurIPS}, 2020.

\end{thebibliography}
}


\newpage

\appendix


\section{Further Details on Related Methods}

Our work is inspired by two SSL methods: \emph{meta pseudo labels (MPL)}~\cite{SSL:MPL} and \emph{co-training}~\cite{SSL:co-training}. We provide more information on these methods below.
This discussion should provide further insight for Algorithm~\ref{alg:MCT}.

\ourparagraph{Meta Pseudo Labels}\label{sec:background:MPL}
In order to discuss 
Meta Pseudo Labels, first we need to discuss Pseudo Labels.
Pseudo Labels (PL)~\cite{SSL:ST:PL} is a broad paradigm for SSL, which involves a teacher and a student in the learning process. The teacher looking at the unlabeled set $U$, selects a few instances and assigns \emph{pseudo-labels} to them.  Subsequently these \emph{pseudo-examples} are shown to the student as if they were legitimate examples so that the student can learn a better model.
In this context, the teacher $f_{\theta_T}$ is fixed and the student is trying to learn a better model by aligning its predictions to pseudo-labeled batches $(X_u, f_{\theta_T}(X_u))$.
Thus, 
on a pseudo-labeled batch, the student optimizes its parameters using 
\begin{equation}\label{eq:PL}
\theta_{S}^{\text{PL}} \in 
\argmin_{\theta_S}\ell\left(\argmax_{\xi} f_{\theta_T}(X_u)|_{\xi}, f_{\theta_S}(X_u)\right)\,.
\end{equation}
The hope is that updated student parameters $\theta_S^{\text{PL}}$ will behave well on the labeled data $L = (X_L, Y_L) = \left\{(x_i, y_i)\right\}_{i=1}^{m}$; 
that is, 
the loss 
$\mathcal{L}_L\left(\theta_{S}^{\text{PL}}\left(\theta_T\right)\right) = 
\ell\left( Y_L, f_{\theta_S^{\text{PL}}}(X_L)\right)$ is small, where 
$\mathcal{L}\left(\theta_{S}^{\text{PL}}\left(\theta_T\right)\right)$ 
is defined 
so that the dependence between $\theta_S$ and $\theta_T$ is explicit.

Meta Pseudo Labels (MPL)~\cite{SSL:MPL} is exploiting this dependence between $\theta_S$ and $\theta_T$.
In other words, because the loss that the student suffers on $\mathcal{L}_L\left(\theta_{S}^{\text{PL}}\left(\theta_T\right)\right)$ is a function of $\theta_{T}$, 
and making a similar notational convention for the unlabeled batch $\mathcal{L}_u\left(\theta_T, \theta_S\right) = \ell\left(\argmax_{\xi} f_{\theta_T}(X_u)|_{\xi}, f_{\theta_S}(X_u)\right)$,
then one can further optimize $\theta_T$ as a function of the performance of $\theta_S$:
\begin{equation}
\left\{
\begin{array}{rl}
\displaystyle\min_{\theta_T} & \displaystyle\mathcal{L}_L\left(\theta_{S}^{\text{PL}}\left(\theta_T\right)\right)\,,\\[0.3em]
\text{where} & 
\displaystyle\theta_{S}^{\text{PL}}(\theta_T) \in \argmin_{\theta_S} \mathcal{L}_u\left(\theta_T, \theta_S\right)\,.
\end{array}\right.
\end{equation}
How we might exploit the dependency between $\theta_T$ and $\theta_S$ might not be immediately obvious. 
Eventually the models are optimized using $\theta_{S}^{\text{PL}} \approx \theta_S - \eta_S\cdot\nabla_{\theta_S} \mathcal{L}_u\left(\theta_T, \theta_S\right)$
as a 
practical approximation~\cite{SSL:MPL:Approximation1,SSL:MPL:Approximation2}
which 
leads to the tractable teacher objective:
\begin{equation}\label{eq:MPL:teacher}
\min_{\theta_T} \mathcal{L}_L\left(\theta_S - \eta_S\cdot\nabla_{\theta_S} \mathcal{L}_u\left(\theta_T, \theta_S\right)\right)\,.
\end{equation}
For a more detailed explanation of this relationship we refer the interested reader to~\cite{SSL:MPL}.


\ourparagraph{Co-Training}\label{sec:background:CT}
Co-training relies on the existence of two different \emph{views} of the instance space $\XX$; i.e., $\XX = \XX^{(1)} \times \XX^{(2)}$. Hence, an instance $x\in\XX$ has the form $x = (x^{(1)}, x^{(2)})\in\XX^{(1)} \times \XX^{(2)}$. 
We 
call
the views $x^{(1)}$ and $x^{(2)}$ 
\emph{complementary}.
Co-training requires two assumptions in order to work.
First, it is assumed that, provided enough many examples, each separate view is enough for learning a meaningful model. 
The second is the \emph{conditional independence assumption}~\cite{SSL:co-training}, which states that $x^{(1)}$ and $x^{(2)}$ are conditionally independent given the label of the instance $x = (x^{(1)}, x^{(2)})$.
%
Thus, with co-training, 
two models $f^{(1)}$ and $f^{(2)}$ are learnt, 
which help each other using different information 
captured from the different views of each instance. 
Below we provide more information.


Given a dataset $S = L\cup U$ (Section~\ref{sec:background} has notation),
we can have access to two different views of the dataset due to the natural partitioning of the instance space $\XX$; that is, 
$S^{(1)} = L^{(1)} \cup U^{(1)}$ and $S^{(2)} = L^{(2)} \cup U^{(2)}$.
Initially two models $f^{(1)}$ and $f^{(2)}$ are learnt using a supervised learning algorithm based on the labeled sets $L^{(1)}$ and $L^{(2)}$ respectively.
Learning proceeds in \emph{iterations}, where in each 
iteration a number of unlabeled instances are integrated with the labeled set by assigning pseudo-labels to them that are predicted by $f^{(1)}$ and $f^{(2)}$, so that improved models $f^{(1)}$ and $f^{(2)}$ can be learnt by retraining on the \emph{augmented} labeled sets. In particular, in each iteration, each model $f^{(v)}$ predicts a label for each view $x^{(v)}\in U^{(v)}$. 
The top-$k$ most confident predictions by model $f^{(v)}$ are used to provide a pseudo-label for the respective \emph{complementary} instances and then these pseudo-labeled examples are integrated in to the labeled set for the complementary view.  In the next iteration a new model is trained based on the augmented labeled set.
Both views of the used instances are dropped from $U$ even if only one of the views may be used for augmenting the appropriate labeled set.
This process repeats until the unlabeled sets $U^{(v)}$ are exhausted, yielding models $f^{(1)}$ and $f^{(2)}$ corresponding to the two views.  Co-Training is evaluated using the \emph{joint prediction} of these models which is the re-normalized element-wise product of the two predictions:
\begin{equation}\label{eq:co-training:joint}
\frac{f^{(1)}\left(x^{(1)}\right)
\otimes
f^{(2)}\left(x^{(2)}\right)}{
\norm{f^{(1)}\left(x^{(1)}\right)
\otimes
f^{(2)}\left(x^{(2)}\right)}_1
}\,.
\end{equation}
where $\otimes$ is the Hadamard (element-wise) product of two vectors; that is, 
for vectors $\alpha, \beta\in\RR^d$ 
their Hadamard (element-wise) product 
is the vector 
$\gamma = \alpha\otimes\beta$, where $\gamma_j = \alpha_j\cdot\beta_j$ for $j\in\set{1, \ldots, d}$.

%

\section{Hyperparameter Configuration}\label{sec:appendix:hyperparameters}
The relevant hyperparameters for training are included in Table~\ref{tbl:hparam-config} for co-training (CT), meta co-training (MCT), and deep ensembles.
\begin{table*}[ht]
\caption{Hyperparmeter configuration for CT, MCT, and deep ensemble.  Other models used (MLP, linear) are trained for the same number of steps as MCT, with the same optimization parameters.}\label{tbl:hparam-config}
    \centering
    \begin{tabular}{cccc}
        \toprule
         config & Co-Training & Meta Co-Training & Deep Ensemble\\    
         \midrule
         optimizer & Adam & Adam & Adam\\
         optimizer momentum & $\beta_1, \beta_2 = 0.9, 0.999$ & $\beta_1, \beta_2 = 0.9, 0.999$ & $\beta_1, \beta_2 = 0.9, 0.999$\\
         base learning rate & 1e-4 & 1e-4 & 1e-4\\
         batch size & 4096 & 4096 & 4096 \\
         learning rate schedule & Reduce On Plateau & Reduce On Plateau & Reduce On Plateau\\
         schedule patience & 10 steps & 10 steps & 10 steps \\
         decay factor & 0.5 & 0.5 & 0.5 \\
         training steps & 200 (per iteration) & 1000 & 1000 (per model) \\
         warmup steps & n/a & 200 & n/a \\
         \bottomrule
    \end{tabular}
\end{table*}

\section{Omitted Experimental Analysis on ImageNet}
In this section we provide some figures illustrating the experimental results on ImageNet 10\% and ImageNet 1\% that we did not include into the main text due to space limitations.

\begin{figure}[ht]
    \centering
    \includegraphics[width=.45\textwidth]{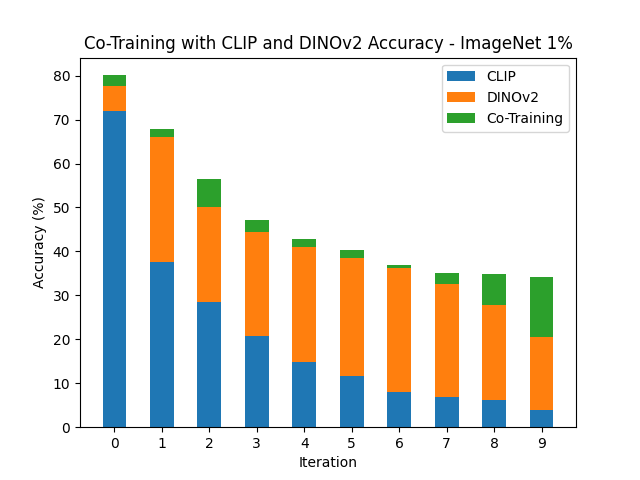}
    \caption{Top-1 accuracy of CT iterations on the CLIP and DINOv2 views for the ImageNet 1\% dataset.\\}
    \label{fig:CD-1}
\end{figure}

\begin{figure}[ht]
    \centering
    \includegraphics[width=0.4\textwidth]{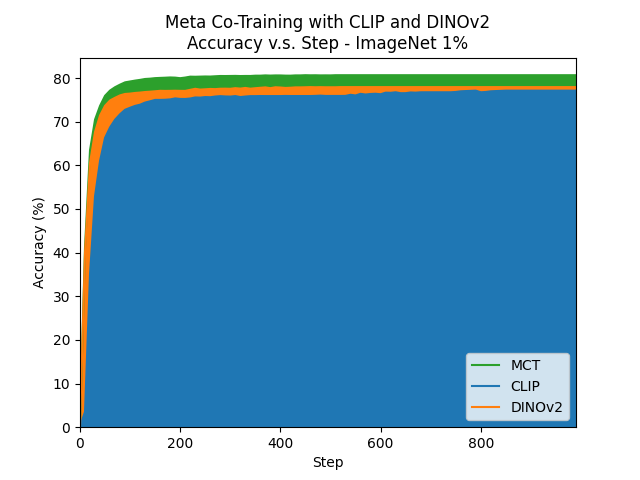}
    \caption{MCT using the CLIP and DINOv2 views as a function of the training step.  Models are trained on 1\% of the ImageNet labels.\\}
    \label{fig:MCTCD1}
\end{figure}

In Figures~\ref{fig:CD-1} and~\ref{fig:MCTCD1} are shown results comparable to those in Figures~\ref{fig:CD-10} and~\ref{fig:MCTCD10} but for the ImageNet-1\% dataset.  
As a reminder these figures are showing top-1 accuracy for CT and MCT when the constituent views used in the embeddings are CLIP and DINOv2.
Similar to the case of ImageNet 10\% we see that the CT performance decreases after the supervised warmup.  Unlike CT, MCT performance increases after the supervised warmup.

\begin{figure}[ht]
    \centering
    \includegraphics[width=.45\textwidth]{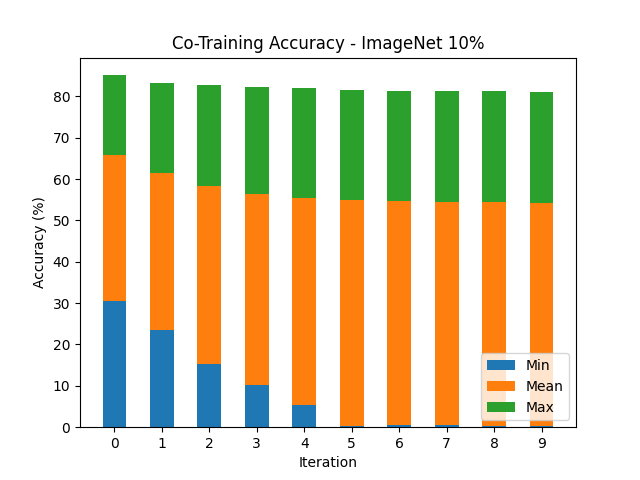}
    \caption{Top-1 accuracy of CT iterations for the ImageNet 10\% dataset. The average over all 10 pairs, the best pair, and the worst pair at each iteration are shown.\\}
    \label{fig:cot-degen-10}
\end{figure}

\begin{figure}[ht]
    \centering
    \includegraphics[width=0.4\textwidth]{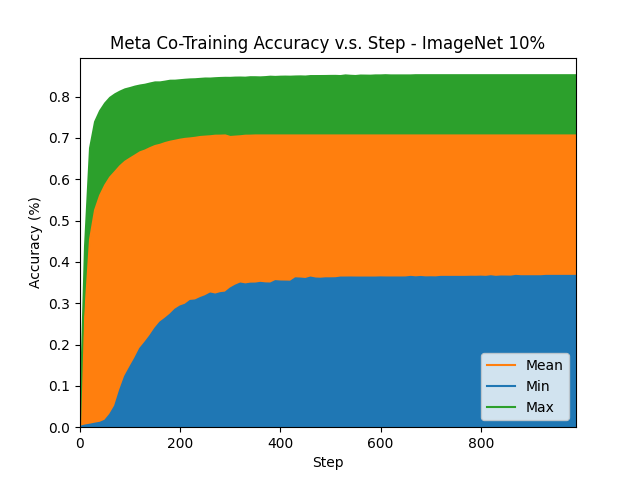}
    \caption{Aggregate statistics of MCT across all views.  Models are trained on 10\% of the ImageNet labels.  The maximum performance of any pair, the minimum performance of any pair, and the average of all 10 pairs is shown.\\}
    \label{fig:MCTMMM10}
\end{figure}

\begin{figure}[ht]
    \centering
    \includegraphics[width=.45\textwidth]{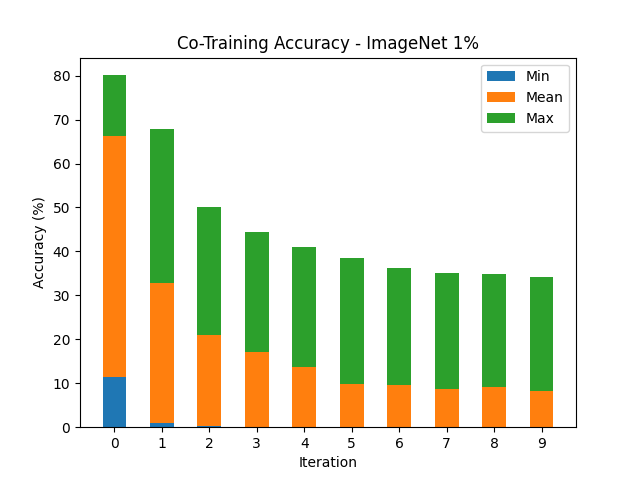}
    \caption{Top-1 accuracy of CT iterations for the ImageNet 1\% dataset. The average over all 10 pairs, the best pair, and the worst pair at each iteration are shown.\\ \\}
    \label{fig:cot-degen-1}
\end{figure}

\begin{figure}[ht]
    \centering
    \includegraphics[width=0.4\textwidth]{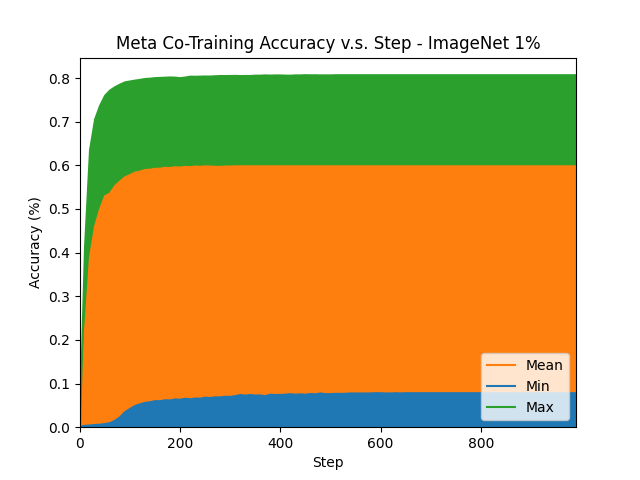}
    \caption{Aggregate statistics of MCT across all views.  Models are trained on 1\% of the ImageNet labels.  The maximum performance of any pair, the minimum performance of any pair, and the average of all 10 pairs is shown.\\ \\}
    \label{fig:MCTMMM1}
\end{figure}
 
In Figures~\ref{fig:cot-degen-10} and~\ref{fig:MCTMMM10} we show the min, max, and mean performance of all of the ten pairs of views for the ImageNet-10\% dataset.  Note that these are not three runs, but aggregate statistics over all runs, so the minimum and maximum points on the graph for one step may not correspond to the same pair as another step.  Figures~\ref{fig:cot-degen-1}  and~\ref{fig:MCTMMM1} are comparable to Figures~\ref{fig:cot-degen-10} and~\ref{fig:MCTMMM10}, but for the ImageNet-1\% dataset.

\section{Further Details on Co-Training Iterations on Additional Datasets}\label{sec:appendix:CT}
In this section we provide further details on the performance of pure CT on additional datasets, using the two views that appear to have the highest information content among those that we tried; i.e., CLIP and DINOv2.  
These experiments support our hypothesis that the primary cause of the failure of CT is 
that one or more views can be weak and do not suffice to learn a model that achieves reasonably high accuracy. 
The closer to equivalent in performance individual views are, the less the performance suffers. When the two views perform similarly and well, the performance improves. 

\begin{figure}[t]
    \centering
    \includegraphics[width=0.45\textwidth]{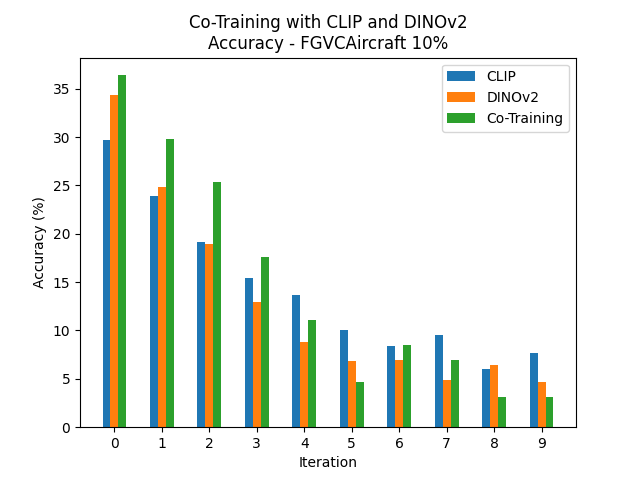}
    \caption{The top-1 accuracy of the CT predictions for each iteration of CT.  10\% of available FGVC Aircraft labels are used for training.  The method exhibits performance improvement over multiple iterations of pseudo-labeling and retraining.}
    \label{fig:CTFGVC10}
\end{figure}

\begin{figure}[t]
    \centering
    \includegraphics[width=0.45\textwidth]{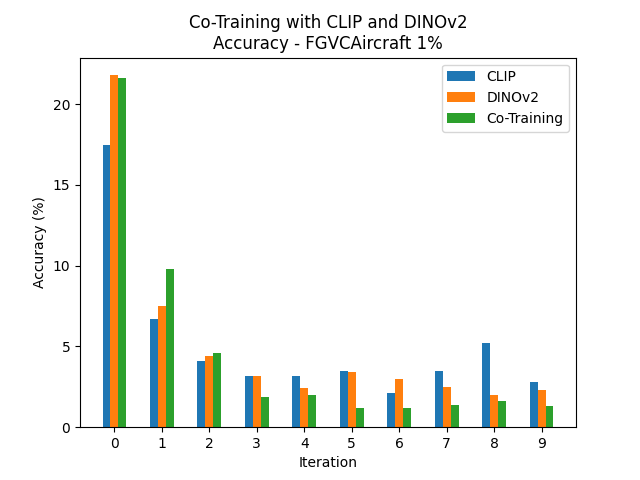}
    \caption{The top-1 accuracy of the CT predictions for each iteration of CT.  1\% of available FGVC Aircraft labels are used for training.}
    \label{fig:CTFGVC1}
\end{figure}

Figure~\ref{fig:CTFGVC10} and Figure~\ref{fig:CTFGVC1} show CT iteration performance on the FGVC Aircraft dataset using 10\% and 1\% of labels respectively.
Neither view performs well and there is a large difference between the performance of each view.  The performance of CT quickly collapses.  

\begin{figure}[ht]
    \centering
    \includegraphics[width=0.45\textwidth]{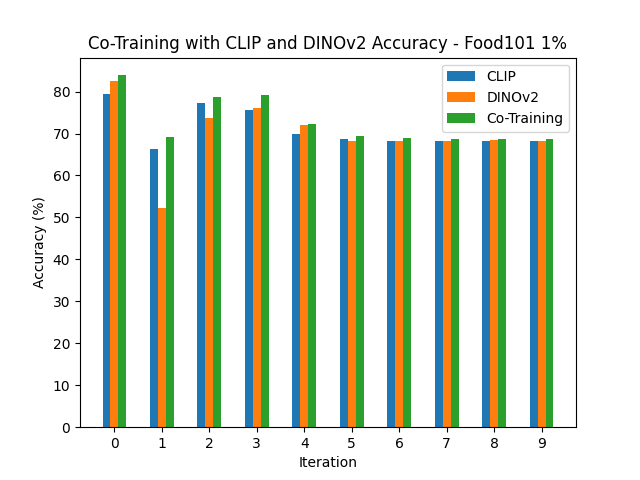}
    \caption{The top-1 accuracy of the CT predictions for each iteration of CT.  1\% of available Food101 labels are used for training.}
    \label{fig:CTFood1}
\end{figure}

Figure~\ref{fig:CTFood10}, which was already presented in the main body of this work, provides an example where CT performs as expected and accuracy increases as CT progresses, though it can be seen from Figure~\ref{fig:CTFood1} that the degeneration still occurs in the case with less labels.

\begin{figure}[t]
    \centering
    \includegraphics[width=0.45\textwidth]{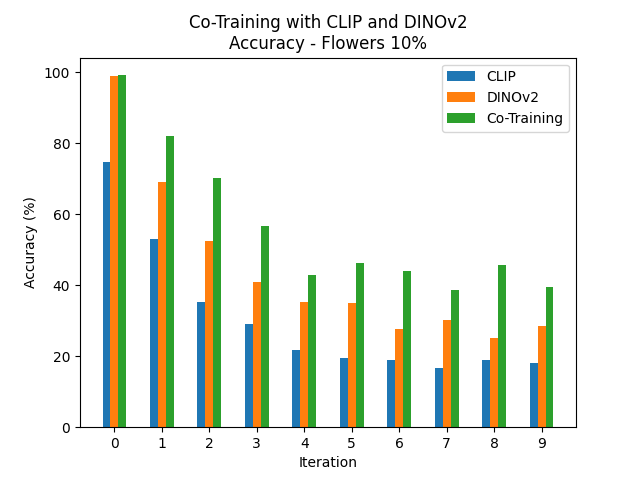}
    \caption{The top-1 accuracy of the CT predictions for each iteration of CT.  10\% of available Flowers102 labels are used for training.}
    \label{fig:CTFlowers10}
\end{figure}

\begin{figure}[t]
    \centering
    \includegraphics[width=0.45\textwidth]{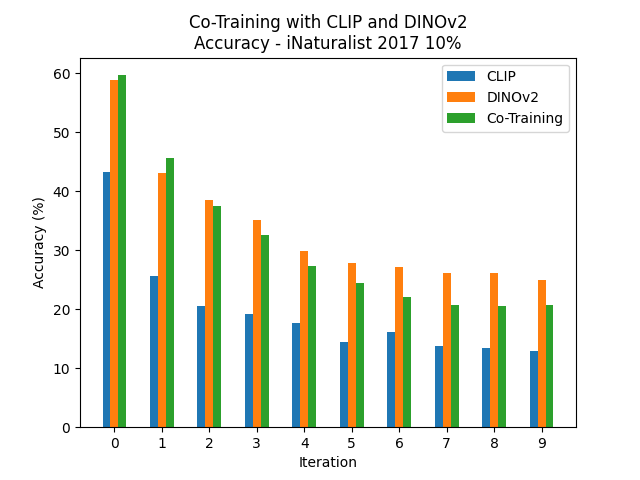}
    \caption{The top-1 accuracy of the CT predictions for each iteration of CT.  Only the 1010 most frequent classes are included.  10\% of available iNaturalist labels from those classes are used for training.}
    \label{fig:CTiNat201710}
\end{figure}

\begin{figure}[t]
    \centering
    \includegraphics[width=0.45\textwidth]{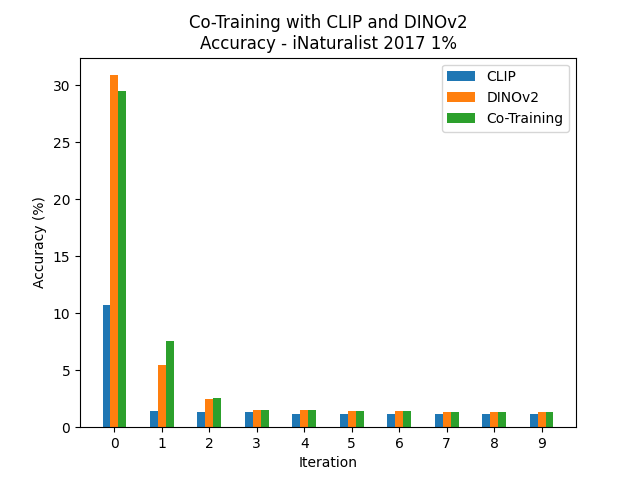}
    \caption{The top-1 accuracy of the CT predictions for each iteration of CT.  Only the 1010 most frequent classes are included.  1\% of available iNaturalist labels from those classes are used for training.}
    \label{fig:CTiNat20171}
\end{figure}

In Figure \ref{fig:CTFood1} we show the iterations of CT for Food101-1\%.  Figure \ref{fig:CTFood10} with the iterations for Food101-10\% appears in \ref{sec:experiments}.  The 1-shot results for Flowers-102 (10\%) appear in Figure \ref{fig:CTFlowers10}.  The iNaturalist-2017 results for 10\% and 1\% respectively appear in Figures \ref{fig:CTiNat201710}, \ref{fig:CTiNat20171}.


\paragraph{Main Takeaway.}
The main takeaway of these additional experiments, apart from establishing new SoTA results in datasets beyond ImageNet, is that 
MCT is in general more robust to view imbalance and provides better results than CT.

\end{document}